# Gated recurrent neural network with TPE Bayesian optimization for enhancing stock index prediction accuracy

Bivas Dinda

Department of Mathematics, Mahishamuri Ramkrishna Vidyapith, Howrah, WB 711401, India.
Email: rs.bivas@knu.ac.in, bvsdinda@gmail.com

**ABSTRACT**

The recent advancement of deep learning architectures, neural networks, and the combination of abundant financial data and powerful computers are transforming finance, leading us to develop an advanced method for predicting future stock prices. However, the accessibility of investment and trading at everyone's fingertips made the stock markets increasingly intricate and prone to volatility. The increased complexity and volatility of the stock market have driven demand for more models, which would effectively capture high volatility and non-linear behavior of the different stock prices. This study explored gated recurrent neural network (GRNN) algorithms such as LSTM (long short-term memory), GRU (gated recurrent unit), and hybrid models like GRU-LSTM, LSTM-GRU, with Tree-structured Parzen Estimator (TPE) Bayesian optimization for hyperparameter optimization (TPE-GRNN). The aim is to improve the prediction accuracy of the next day's closing price of the NIFTY 50 index, a prominent Indian stock market index, using TPE-GRNN. A combination of eight influential factors is carefully chosen from fundamental stock data, technical indicators, crude oil price, and macroeconomic data to train the models for capturing the changes in the price of the index with the factors of the broader economy. Single-layer and multi-layer TPE-GRNN models have been developed. The models' performance is evaluated using standard matrices like $R^2$, MAPE, and RMSE. The analysis of models' performance reveals the impact of feature selection and hyperparameter optimization (HPO) in enhancing stock index price prediction accuracy. The results show that the MAPE of our proposed TPE-LSTM method is the lowest (best) with respect to all the previous models for stock index price prediction.

1. Introduction

The stock markets are complex systems where traders and investors can participate in publicly traded companies for buying and selling shares, hoping to profit from price fluctuations. Access to trading and investment at the fingertips of every interested person is increasingly making the stock market further intricate, noisy, and unpredictable. All the stock market participants are asking for higher profits and lower risk. Stock price or stock index price prediction is undeniably the upper choice for investors and financial analysts due to its extensive implementation area and significant effect on profit and loss. The future stock price or stock index price forecast is multifaceted and perplexing as financial time series are commonly nonlinear, dynamic, and highly volatile (Rezaei et al., 2021; Yujun et al., 2021; Xu et al., 2022).

The possible causes of uncertain stock index price fluctuations are macroeconomic data, political instability, monetary policy of influencing countries, geo-political tension, energy price volatility, public sentiment about the economy, inflation, performance, and management integrity of the stocks traded in the exchange, financial news, etc. To forecast the near future stock index price, traders use different strategies, including technical analysis that shows trends in the stock prices of the index and fundamental analysis that tests the financial conditions and management integrity of companies and their industry condition traded in that index (Hoseinzade & Haratizadeh, 2019; Nelson et al., 2017; Persio & Honchar, 2016; Qiu & Song, 2016; Kara et al., 2011). In recent years, investors, traders, financial advisors, influencers, etc., have shared their thoughts on stocks on various social media platforms, like Facebook posts, tweets on X, YouTube videos, and many other social media platforms. Social media posts are most likely to increase further in the future. Although the information presented to the market participants through social media is incomplete and sometimes intentional, it increases or decreases the sentiment about particular stocks. Recent research shows a positive correlation between sentiment and stock price.

A suitable model formation with the best feasible features is challenging. One probable reason for not attaining the desired output could be not selecting the appropriate features. The features in this study are chosen very carefully by viewing the several sides of the economy and their possible impact on the targeted price of the particular stock or stock index. By considering a well-rounded combination of features, a model can capture a more balanced understanding of the data, leading to improved performance.

Machine learning models have emerged as a promising financial time series prediction approach, demonstrating improved accuracy compared to traditional techniques. Deep learning has been verified as the best-performing machine learning architecture in recent years. Deep learning algorithms can learn from and analyze large quantities of data to create forecasts based on this learning and analysis. Deep learning models can help to make better investment decisions, resulting in increased financial returns.

Recent developments in deep learning architectures can solve different data science problems. However, most architecture treats each data point independently, discarding information from previous steps, resulting in inefficiencies in processing financial time series data. During the training process of recurrent neural network (RNN) models, the inaccuracy is measured by comparing the forecasted values with the real values of the training dataset. The errors are then reduced by updating the bias and weights at each step until the lowest possible error is achieved. The learning process involves using a gradient, which computes the error rate for each parameter. The gradient then minimizes errors by stochastically adjusting the weights and biases. This process is called back-propagation, where the error propagates backward to the input level. One of the main limitations of this network is that gradients can disappear for the parameters of the earlier layers of the network. GRNNs are typical RNN architecture constructed to get over the vanishing gradient problems.

Retaining relevant information for more extended periods is the default behavior of the LSTM and GRU models.

GRNN models, like other deep learning models, require optimized hyperparameters (like learning rate, number of units in each layer, batch size, optimizer, number of layers, etc.) to train models. A proper selection of hyperparameters can significantly enhance the representation of features and models' performance. However, HPO is a challenging assignment. Classical hyperparameter optimization procedures are grid search, random search, genetic algorithm, and Bayesian optimization (BO). BO requires fewer computational resources than a genetic algorithm. BO has two executions: Tree-structured Parzen estimator (TPE) and Gaussian Process (GP). TPE has proven more accurate than GP since TPE scales better to high dimensional spaces and can handle categorical and conditional parameters (Bergstra et al., 2011). Therefore, we choose TPE Bayesian optimization for HPO to select the best hyperparameters for our proposed GRNN models.

The *objective* of this study is to increase the prediction accuracy of the future price of the NIFTY 50 index using TPE-GRNN. The study inputs eight cautiously chosen features from macroeconomic data, fundamental data, and technical indicators to capture the NIFTY 50 index price variation with macroeconomic price variation. When achieving the goal, we comprehensively compare the predictive performance of LSTM, GRU, GRU-LSTM, and LSTM-GRU models by the error matrices RMSE, MAPE, and $R^2$. We intend to optimize the hyperparameters such as learning rate, number of neurons per layer, batch sizes, number of layers, number of epochs, and optimizers with TPE Bayesian optimization for HPO to improve the models' performance.

This study focuses on a computational framework that aims to predict the next day's closing price of the NIFTY 50 index using LSTM, GRU, GRU-LSTM, and LSTM-GRU models. The research framework's schematic diagram is shown in Fig. 3 to achieve the goal. The dataset is cleaned after carefully selecting features and collecting data from Yahoo Finance, and Investing.com. Then, the dataset is normalized using the min-max data normalization technique. The normalized dataset is now ready to be executed by LSTM, GRU, hybrid GRU-LSTM, and LSTM-GRU models. TPE Bayesian optimization for the HPO technique is now applied. Once the optimum hyperparameters are found, the normalized dataset is executed in GRNN models with optimum hyperparameters. The prediction accuracy of the proposed models is evaluated using the error metrics of $R^2$, MAPE, and RMSE.

The following are the main contributions of this study.

1. Carefully choosing the input features by observing different aspects of the economy (stock market index is a trustworthy indicator of the overall economy's strength and momentum) and combining them into a unified structure to form reliable GRNN models for predicting stock index prices with highest accuracy.
2. TPE Bayesian optimization for HPO is conducted effectively, searching for optimal hyperparameters for the best prediction accuracy of GRNN models.

3. Different GRNN models such as LSTM, GRU, GRU-LSTM, and LSTM-GRU are assimilated for best fit and low error under identical conditions.
4. Performing statistical experiments to validate the prediction accuracy of considered GRNN models.

The paper is structured as follows. Section 2 discusses related works. The GRNN model's implementation methodology, modeling framework, and research design are explained in section 3. The feature selection process, data collection, and input preparation are presented in section 4. The experimental setup, hyperparameter tuning for GRNN models, model result comparison, and experimental analysis are presented in section 5. Section 6 presents conclusions.

2. **Related work**

Many researchers tend to focus on predicting stock market indices instead of predicting the price of a single stock. This is because indices are usually less unstable than a particular stock, consisting of different stocks from various sectors, providing a broader representation of the economy's pace and general state.

In the literature, several research studies have been conducted on predicting stock price or stock index price using LSTM and GRU. Some of them are listed below. Chen et al. implemented LSTM to forecast the CSI 300 (China) index returns, achieving 27.2% accuracy (Chen et al., 2015). Bao et al. combined wavelet transform, stacked auto-encoders, and LSTM used on S&P 500 data (Bao et al., 2017). Their best (lowest) MAPE is 0.024%. Roondiwala et al. applied LSTM to NIFTY 50 data using OHLC data from 2011 to 2016; their lowest RMSE on normalized data is 0.00859 (Roondiwala et al., 2017). Fischer and Krauss successfully utilized LSTM for directional movement prediction of the stocks contained in the S&P 500 from 1992 to 2015 (Fischer & Krauss, 2018). Karmiani et al. compared LSTM, SVM, Backpropagation, and Kalman filter (Karmiani et al., 2019). LSTM gives the best prediction accuracy. Yu and Yan compared LSTM, MLP, SVM, and ARIMA models worldwide across index data (Yu & Yan, 2019). LSTM outperformed other model architectures. Qiu et al. used attention-based LSTM on the Hang Seng index (HSI), S&P 500, and Dow Jones (DJIA) (Qiu et al., 2020). They employed wavelet transformation to de-noise the dataset and used news data attention to predict the opening price of the said indexes using fundamental data. LSTM is used for high-frequency trading of 1, 5, and 10 minutes to predict S&P 500 prices by Lanbouri and Achchab (Lanbouri & Achchab, 2020).
Yadav et al. used LSTM models with several layers of stock data from India and concluded that single-layer LSTM provides the best accuracy in closing price prediction (Yadav et al., 2020). Bhandari et al. used LSTM in the S&P 500 dataset with multiple features and a manual search for HPO to predict the next-day closing price of the S&P 500 index (Bhandari et al., 2022).

Di Persio and Honchar compared the performance of various RNN models on Google stock, such as basic RNN, LSTM, and GRU. They observed the outperformance of the LSTM network (Di Persio &

Honchar, 2017). Pokhrel et al. analyzed the performance of CNN, LSTM, and GRU using NEPSE (Nepal) data and several influencing factors with a manual search of hyperparameters (Pokhrel et al., 2022). They find the LSTM to be the best predictive model among others. Gülmez used an artificial rabbits algorithm(ARO) and genetic algorithm (GA) as HPO and optimized LSTM for stock price prediction (Gülmez, 2023). The results show that LSTM-ARO outperforms all other models with best MAPE =0.020% and MSE=1.905 for DJIJ (DOW) data. It is seen that single-layer LSTM models produce better prediction accuracy than higher-layer LSTM models (Yadav et al., 2020; Bhandari et al., 2022; Pokhrel et al., 2022; Gülmez, 2023). Beniwal et al. forecast multistep daily stock index price of five significant global indices and show that LSTM performs better with MAPE=3.95% and RMSE=881.38 for the Nifty 50 dataset (Beniwal et al., 2024).

Deng et al. developed a MEMD-LSTM model with an orthogonal array tuning method (OATM) for HPO to predict multi-step ahead stock price and showed the model's outperformance with MAPE=0.5276% and RMSE=36.0193 for S&P 500 index data (Deng et al., 2022). Some studies have been done on the hybridization technique of RNN models. Hossain et al. created a hybrid LSTM-GRU deep learning model using the S&P 500 index data of 66 years (1950 to 2016) (Hossain et al., 2018). Islam and Hossain predicted Foreign exchange currency rates using a hybrid GRU-LSTM model (Islam & Hossain, 2021).

This review of research suggests that there's a rise in research on combining different models (hybridization technique) to forecast stock prices. These hybrid models possess entirely new techniques that merge various theoretical models. Finally, the study indicates the importance of factors like the chosen input data, different input factors, especially the HPO methods, and the architecture of the deep learning model itself, as these can significantly impact the prediction accuracy.

## 3. Methodology

The proposed modeling approach uses four different gated deep learning algorithms, LSTM, GRU, GRU-LSTM, and LSTM-GRU, to predict NIFTY 50 (India) index price. For sequential time series data, these GRNN models can produce superior results. We aim to predict the NIFTY 50 index price using those GRNN models based on a multi-factor dataset of historical prices, technical indicators, and macroeconomic data.

### 3.1 LSTM model

LSTM networks are a particular kind of gated RNN that excels at learning long-term dependencies (or solves short-term memory issues of RNN) in sequential data (Hochreiter & Schmidhuber, 1997). The LSTM was proposed in 1997, and over the years, several researchers have worked to improve it gradually (Gers et al., 2000; Gers et al., 2003; Sak et al., 2014).

LSTM units are composed of memory cells with input, output, and forget gates, as shown in Fig. 1. Each memory cell has the same input and output process as a simple RNN. Also, each cell has a gating unit that controls information flow called forget gate $f_t$ (for time step t).

$$f_t = \sigma(b_f + V_f x_t + W_f h_{t-1})$$

Where $x_t$ = current input vector, $h_t$ = current hidden layer vector, $h_t$ contains the outputs of previous LSTM cells.

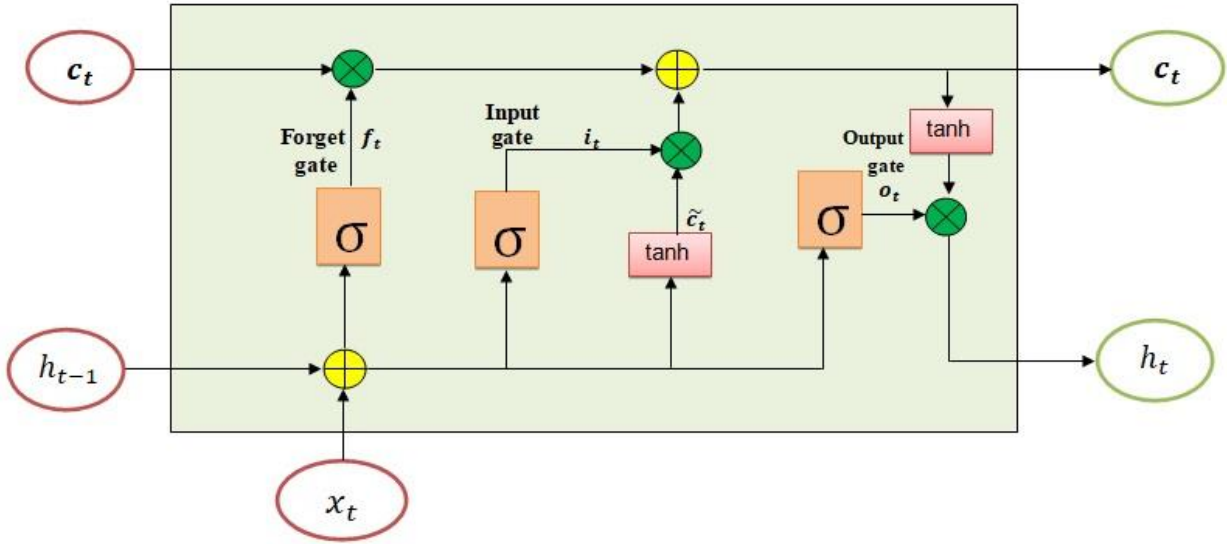

**Fig. 1.** A memory cell of LSTM architecture

The cell state unit $c_t$ stores information in memory cell:

$$c_t = f_t \otimes c_{t-1} + i_t \otimes \tilde{c}_t$$
$$\tilde{c}_t = \tanh(b_c + V_c x_t + W_c h_{t-1})$$

The input gate $i_t$ regulates the information flow into memory cell:

$$i_t = \sigma(b_i + V_i x_t + W_i h_{t-1})$$

The output gate regulates the output of memory cell to the remaining network:

$$o_t = \sigma(b_o + V_o x_t + W_o h_{t-1})$$

The hidden state, or the output of the LSTM, makes a prediction or forwards the information to the next LSTM memory cell.

$$h_t = \tanh(c_t) \otimes o_t$$

Where tanh and $\sigma$ are the hyperbolic tangent and sigmoid function, and $b_A, V_A, W_A$ respectively represent biases, input weights, and recurrent weights of the gates $A = \{f = $ forget gate, $i = $ input gate, $o = $ output gate$\}$.

An LSTM cell takes three different inputs: the current input (new information) $x_t$, the short-term information of the previous cell (short-term memory) $h_{t-1}$, and the long-term memory from the preceding cell $c_{t-1}$. The forget gate acts like a sorter for the information of $x_t$ and $h_{t-1}$. Using the sigmoid layer, the forget gate checks the relevance of the information $x_t$ and $h_{t-1}$ by giving them weights between 0 and 1. If the weight is 0, then the cell forgets all the information from the previous cell, and if the weight is 1, then all the information is kept.

*3.2 GRU model*

GRU is a simpler alternative to the LSTM network and performs as well as LSTM (Cho et al., 2014; Chung J. et al., 2014; Chung J. et al., 2015; Jozefowicz et al., 2015; Chrupała et al., 2015; Greff et al., 2016). Unlike LSTM, a single gating unit in GRU controls the storing and forgetting factors concurrently. The GRU has two gating mechanisms: reset and update gates, as presented in Fig. 2.

The reset gate $r_t$ determines the quantity of previous information that should be ignored.

$$r_t = \sigma(b_r + V_r x_t + W_r h_{t-1})$$

The update gate $z_t$ controls the amount of the previous knowledge that needs to be forwarded into the future. The update gate is identical to the input gate and forget gate of LSTM.

$$z_t = \sigma(b_z + V_z x_t + W_z h_{t-1})$$
$$\tilde{h}_t = \tanh(b_c + V_c x_t + W_c (r_t \otimes h_{t-1}))$$
$$h_t = (1 - z_t) \otimes h_{t-1} + z_t \otimes \tilde{h}_t$$

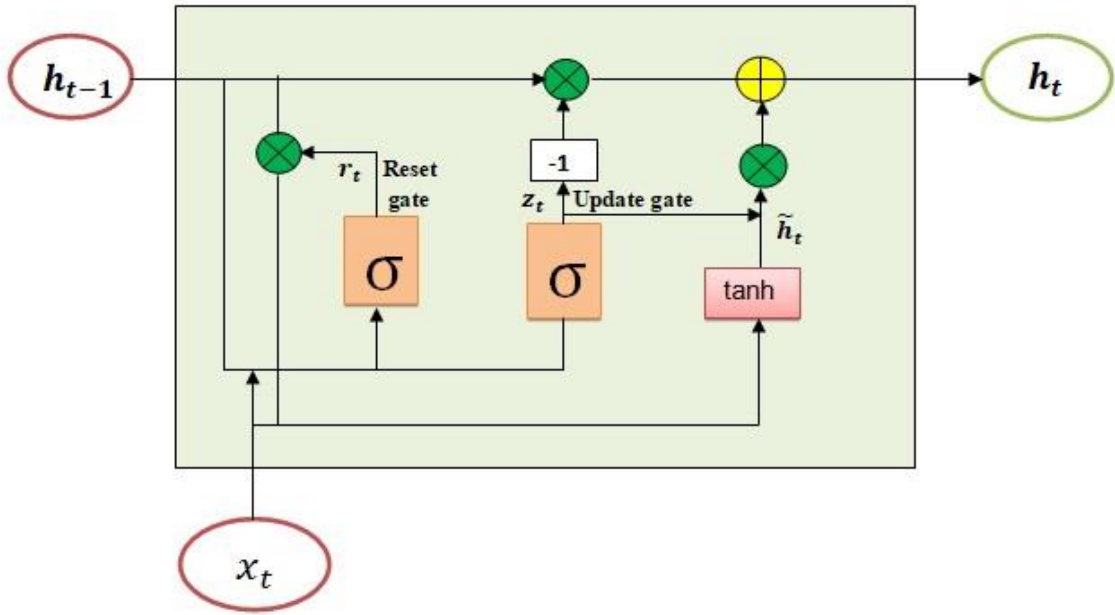

**Fig. 2.** A memory cell of GRU architecture

*3.3 Proposed model framework*

The modeling framework used in this paper is presented by the flow chart revealed in Fig. 3. The data of different features are collected from various sources (Table. 1). Then, the dataset undergoes a data pre-processing or input preparation stage, which includes applying technical indicators, data cleaning, and data normalization. Afterward, different GRNN models are constructed based on the optimum hyperparameters obtained from TPE Bayesian optimization for HPO of each model. Then, we train the models and save the models with the lowest possible error matrices. Then, the accuracy of the prediction will be visualized.

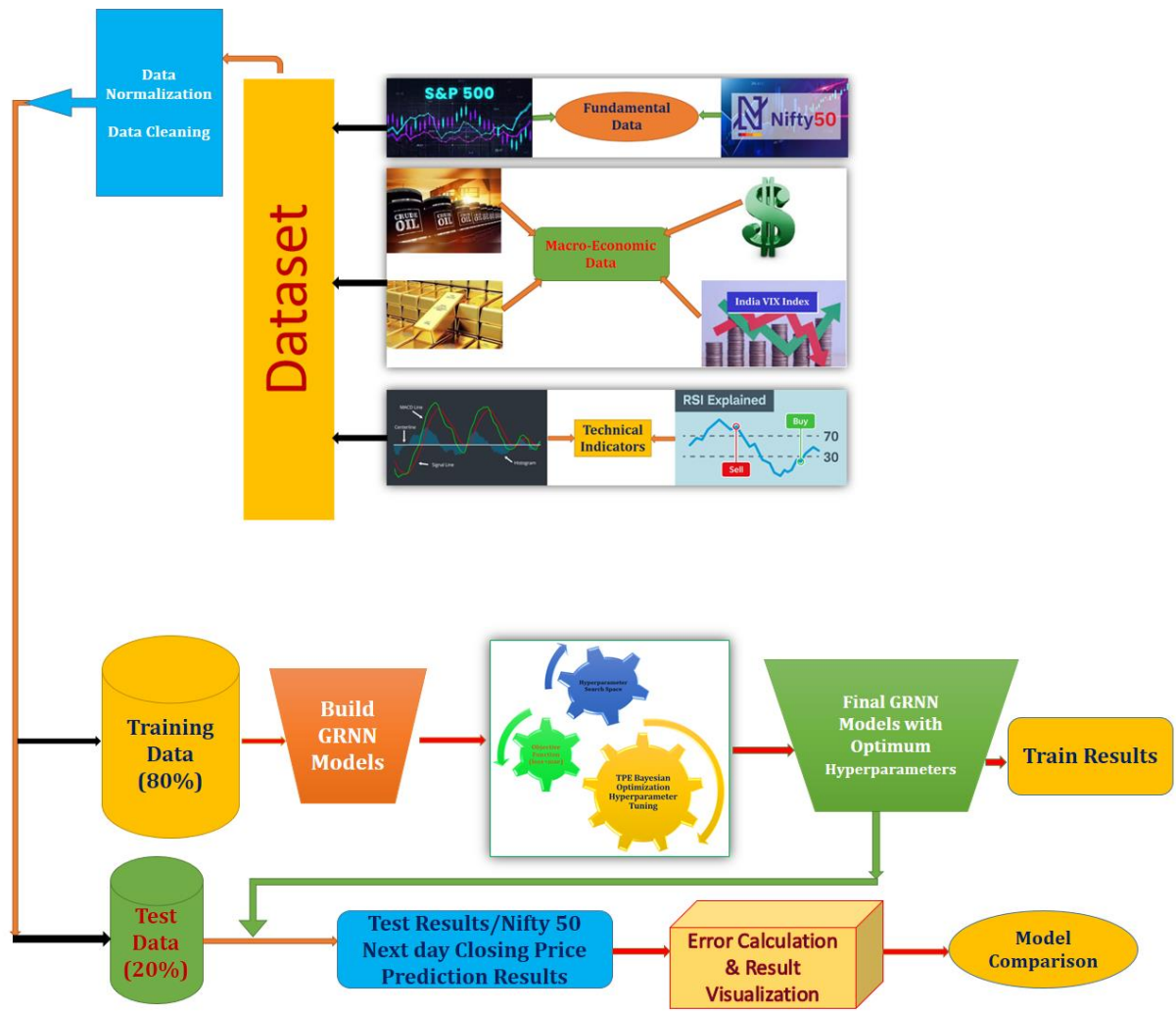

**Fig. 3.** Empirical flow chart of proposed research framework

## 4. Data description

The National Stock Exchange (NSE), a stock exchange in India, is headquartered in Mumbai. NSE is the world's largest derivative exchange in terms of contacts traded, and it is the third largest in the equity segment by the number of traders in 2023, as per statistics mentioned by WFE (World Federation of Exchanges). NIFTY 50, the benchmark index of NSE, executes the price of 50 market-capitated publicly traded company stocks in India. The NIFTY 50 time series data is collected from yahoo finance from 2 January 2009 to 31 December 2023. The fifteen years of NIFTY 50 data contains 3649 samples after calculating MACD and RSI. The 15 years' time period contains both bull and bear markets because of many triggers, some of them are global economic recovery after the 2008 financial crisis (2009-2011), Eurozone debt crisis and global economic slowdown (2011-2013), rising interest rates, and concerns about Chinese economic growth (2015-2016), rising oil prices, trade tensions, and global economic slowdown (2018-2019), COVID-19 pandemic (2020-2021).

**Table 1.** List of features for the model

| Features | Data sources | Frequency | Abbreviations |
| --- | --- | --- | --- |
| NIFTY closing price (in ₹) | *Yahoo Finance* | Daily | NIFTY |
| S & P 500 closing price (in $) | *Yahoo Finance* | Daily | S & P 500 |
| Crude oil price (in $) | *Yahoo Finance* | Daily | Crude Oil |
| NIFTY volatility index (in ₹) | investing.com | Daily | India VIX |
| USD to INR exchange rate (in ₹) | *Yahoo Finance* | Daily | INRUSD |
| Gold price (in $) | *Yahoo Finance* | Daily | Gold |
| MACD | - | Daily | MACD |
| RSI | - | Daily | RSI |

This section briefly describes the strategy for selecting features, normalizing data, and preparing model inputs. The proposed model predicts the following-day closing price of the NIFTY 50 index based on technical indicators, fundamental data, and macroeconomic data. All the input features of our proposed model are presented in Table 1. All the features considered in the projected model contain daily data.

### 4.1 Feature selection strategy

Our consideration's The fundamental or historical data fo our consideration are NIFTY50 and S&P500 index closing price. The S&P 500 and NIFTY 50 are major stock indices that reflect the overall health of their respective markets (India and USA). The Indian and USA economies are somewhat interconnected, and the S&P 500 exchange closes after nine and half hours of the NIFTY 50 index closes. Therefore, major economic news, sudden geopolitical tension, or any adverse events in the US or the world can have ripple effects on S&P 500 closing price, impacting NIFTY50 closing price the next day.

India imports over 80% of their oil needs. A weaker INR (Indian rupee) concerning USD (US dollar) can increase import costs, impacting companies' bottom lines and potentially pulling down the NIFTY50 index. A weakening rupee can create uncertainty and negatively impact investor sentiment, leading to potential selling pressure on stocks and affecting the NIFTY 50. For example, in 2013, the rupee crisis saw the NIFTY50 index fall alongside the weakening rupee, and between the years 2016 and 2020, a stronger rupee coincided with the rising NIFTY 50 index. Therefore, the USD to INR exchange rates significantly impact the NIFTY 50 index's next-day closing price prediction.

Crude oil is a crucial raw material for many businesses; businesses must pay more for their inputs when prices increase. This can cause an increase in prices, leading to a reduction in demand for goods and services (Kilian, 2008; Herrera et al., 2019). This can decrease corporate profits, resulting in a decline in the stock market. Crude oil prices decline sharply, sometimes implying a decline in demand from the countries with the highest oil imports, China and India. This provokes investors to rethink the economic growth of China and India, leading to a decline in the NIFTY 50 index price. The crude oil price significantly impacts the prediction of stock index prices (Park & Ratti, 2008).

Combining technical analysis with fundamental factors can provide a more comprehensive picture of financial time series. MACD (moving average convergence divergence) is a popular and effective technical indicator in stock price momentum analysis. MACD highlights potential trend reversal and changes in momentum. MACD is the connection between the moving averages of a stock price or index price. MACD is calculated by the difference between the 12-day exponential moving average (EMA) and 26-day EMA (Murphy, 1999). When the MACD crosses above zero, it is generally interpreted as a bullish signal and vice versa. Many researchers use MACD to make investment decisions (Chong et al., 2014; Anghel, 2015; Rodríguez-González et al., 2011).

Relative strength index (RSI) is an oscillator technical indicator that measures the recent price changes of a stock and stock index to estimate whether it is overbought (likely to decline) or oversold (likely to rise) (Rodríguez-González et al., 2011). A trade signal is generated when the RSI exceeds the predicted thresholds. When the RSI of a stock or stock index is above 70, the stock or stock index is considered overbought, and when the stock or stock index is below 30, it is considered oversold.

In times of war, pandemic, or other economic uncertainty, investors turn towards a safe haven asset like gold. This can cause the gold price to rise, leading to a decline in the stock market. The gold price can affect investor sentiment. When the gold price goes up, investors may become more doubtful about the economy's future, resulting in a decline in the stock market. In 2008, gold price reached a record high of over $1,030 per ounce, and there was a sharp decline in the Nifty 50 index price. In 2011, the gold price reached another record high of over $1,900 per ounce, and the Nifty 50 index price also declined. In 2020, the gold price reached a new record high of over $2,000 per ounce. This also led to a decline in the Nifty 50 index price. Gold price fluctuation strongly correlates with geopolitical tension (Baur & McDermott, 2010; Selmi et al., 2022; Chiang, 2022). The geopolitical risk or gold price can significantly affect forecasting the future price of NIFTY 50 index (Mishra et al., 2010; Bhunia & Das, 2012; Anand & Madhogaria, 2012).

India VIX is an index used to measure the volatility of the Nifty index based on its option prices. It calculates the expected market volatility percentage (volatility is often referred to as risk in finance) for the next 30 calendar days based on the bid-ask prices of Nifty option contracts. Chicago Board Options Exchange (CBOE) and S&P have granted NSE a license to use the name 'India VIX' to represent the volatility index of Nifty option prices. India VIX and NIFTY index returns have an opposite relationship. Therefore, India VIX can effectively contribute to the future price prediction of the NIFTY 50 index.

*4.2 Data normalization*

Stock index data exhibits volatility and is recorded at fixed intervals. In the dataset, eight different input factors with various scales are chosen to forecast the following-day closing price of the NIFTY 50 index. Data normalization is crucial for a multifactor LSTM or GRU model for several reasons.

- Gradient Descent Optimization**:** LSTM and GRU models use gradient descent to optimize their internal weights and biases. During training, the gradients adjust the weights in a direction that minimizes the loss function. If features have vastly different scales (as shown in Table 2), features with large scales will dominate the gradients, and features with more minor scales might have negligible gradients. Data normalization ensures all the features contribute proportionally to the gradient, allowing for balanced gradient updates and facilitating smoother and more efficient optimization.

- Internal Activation Functions: LSTM models utilize activation functions within their internal layers, like tanh or sigmoid. These functions operate on a specific range of input values. For instance, values outside the range of -1 to 1 for tanh can lead to vanishing gradients, where updates become extremely small and learning stagnates. When features have significantly different scales, some activations might saturate (always outputting the same value) for features with large values, effectively removing their influence. Normalization prevents saturation and ensures the activation functions operate within their optimal range for all features.
- Standardization across Factors: In a multifactor LSTM, different factors might be measured in other units (dollars, rupees, etc.). Normalization puts all aspects on a standard scale, allowing the model to learn the relationships between them more effectively. It also helps the model understand the relative importance of these values for prediction.

Normalization ensures that all factors have a fair shot at influencing the model's learning process, leading to more accurate and robust stock price predictions.

**Table 2.** Statistical description of features data

|      | NIFTY    | Crude Oil | India VIX | INRUSD | Gold    | S&P 500 | RSI   | MACD     |
|------|----------|-----------|-----------|--------|---------|---------|-------|----------|
| Mean | 9735.03  | 71.27     | 19.45     | 64.01  | 1455.14 | 2445.92 | 54.97 | 34.04    |
| Std. | 4515.03  | 21.60     | 7.82      | 11.34  | 290.44  | 1106.28 | 17.30 | 127.05   |
| min  | 2573.15  | 37.63     | 10.14     | 43.90  | 139.60  | 676.53  | 6.98  | -1005.75 |
| max  | 21778.3  | 123.70    | 83.61     | 85.19  | 2081.9  | 4798.30 | 95.02 | 451.08   |

The min-max scaling for normalization of the input features data has been implemented to address the above concerns. In the min-max normalization technique normalized data, $z = \frac{x - x_{min}}{x_{max} - x_{min}}$ where x, $x_{max}$, $x_{min}$ are original inputs, the maximum, and the minimum of the inputs correspondingly. We divided the dataset into 80% for training and 20% for testing. Once the hyperparameter tuning is done, the GRNN models are fitted on the training data with optimal hyperparameters.

## 5. Experiment and results

The data of the selected features has been normalized, and necessary steps have been taken to reshape and split the time series dataset into training (80%) and testing (20%) sets. The objective is to forecast the closing price of the NIFTY 50 index for the following day. Many factors like trend, seasonality, cyclicality, irregularity, and external factors can impact time series. The overall movement of the closing price of NIFTY 50 is upward (as shown in Fig. 4) in spite of various irregularities in price and many external factors.

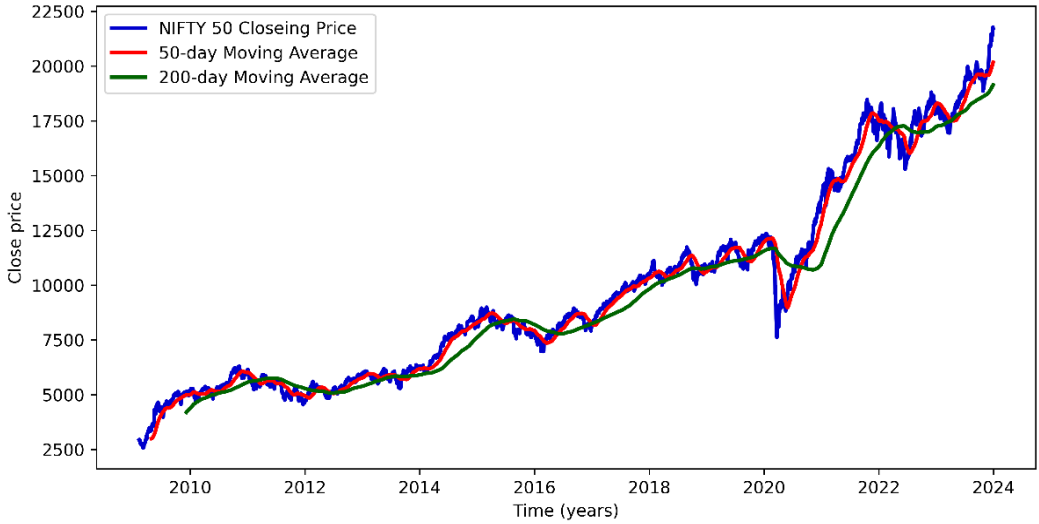

**Fig. 4.** NIFTY 50 closing price along with simple moving averages

*5.1 Evaluation of predictive performance*

Various GRNN algorithms are implemented, including LSTM, GRU, GRU-LSTM, and LSTM-GRU models, to forecast the following-day closing price of the NIFTY50 (India) index. Within each of these models, optimum numbers of neurons, optimum numbers of layers, and optimum learning rates are considered for each model to find the best model. The prediction accuracy and comparison of these models are assessed by calculating $R^2$, RMSE, and MAPE.

$R^2$ score measures how well the model fits with the data. The best $R^2$ score is 1.0. It can be negative if the model fits worse.

$$R^2(y, \hat{y}) = 1 - \frac{\sum_{i=1}^{n}(y_i - \hat{y}_i)^2}{\sum_{i=1}^{n}(y_i - \bar{y})^2}, \qquad \bar{y} = \frac{1}{n}\sum_{i=1}^{n} y_i$$

RMSE can be measured by the dissimilarity concerning actual and predicted values. Lower RMSE implies a better model.

$$RMSE(y, \hat{y}) = \sqrt{\frac{1}{n}\sum_{i=0}^{n-1}(y_i - \hat{y}_i)^2}$$

MAPE is an error metric for regression problems. The idea of this metric is to be sensitive to the mean absolute error. Lower MAPE implies a better model.

$$MAPE(y, \hat{y}) = \frac{1}{n}\sum_{i=0}^{n-1}\frac{|y_i - \hat{y}_i|}{\max(\epsilon, |y_i|)}$$

Where $\epsilon$ is a strictly positive arbitrary small positive real number to dodge undefined results when $y$ is zero, the MAPE function supports multi-output.

$R^2$, MAPE, and RMSE are calculated on the inverse transformation on the predicted normalized data. To remove the biases by the scale of the original data (Table 2), we also calculate RMSE on normalized data (RMSE (ND)). For addressing the stochastic behavior of the weights, each of the models is trained at least 32 times independently. The best possible $R^2$, MAPE, and RMSE obtained from multiple time execution is considered as model selection criteria. The best model is considered by the largest possible $R^2 (\leq 1)$, and the smallest possible MAPE and RMSE of the model.

*5.2 Environment setup*

The development tool used in this study is Jupyter Notebook 7.1.1. The code implementation is based on Python 3.10.11, Tensor Flow 2.16.1, Keras API 3.1.1, pandas 2.2.1, NumPy 1.26.4, sci-kit-learn 1.4.1, scripy 1.12.0, and matplotlib 3.8.3. The processed data is saved locally as a .csv file. The best models for each architecture are saved locally as .keras files. Google Collab GPU is used for TPE Bayesian optimization HPO of the GRNN models.

*5.3 Hyperparameter tuning using TPE Bayesian optimization*

Choosing the right hyperparameters has traditionally been a complex and time-consuming process, often relying on intuition, experimentation, and manual search methods. This is where HPO comes into play. HPO offers three key benefits: 1. It automates the hyperparameter tuning process, saving significant time and resources. 2. By exploring a wider range of hyperparameter combinations, HPO can identify configurations that lead to better model performance. 3. HPO helps find consistent and reliable hyperparameter settings, making it easier to replicate successful model results.

The hyperparameters for the NIFTY 50 index price prediction models using GRNN architectures are the number of layers, the number of units in each layer, batch size, learning rate, optimizers, regularization techniques, activation functions, momentum value, loss function, and number of epochs. Better HPO produces improved prediction accuracy of a model. There are different traditional methods for finding optimum hyperparameters; we use the TPE Bayesian optimization technique for HPO to enhance the models' predictive performance.

The TPE Bayesian optimization HPO includes optimizing the learning rate, number of units in each layer, and batch size. The HPO process includes defining objective functions to minimize errors, considering a hyperparameter search space, and executing a TPE Bayesian optimization algorithm with 60 iterations.

To optimize the other hyperparameters, we will use the following approach. First, we search for an optimum optimizer. AdaGrad (Adaptive Gradient Algorithm) is an improved SGD (stochastic gradient descent) that converges more quickly than SGD (Duchi et al., 2011). RMSProp (Root Mean Square Propagation) optimizer can include the learning rate for each parameter (Tieleman, 2012). ADAM (Adaptive Momentum Estimation) is an improved form of RMSProp. ADAM combines the average of RMSProp and AdaGrad (Kingma & Ba, 2014). NADAM optimizer combines ADAM and Nesterov accelerated gradient (Wilson et al., 2017). NADAM optimizer often converges slightly faster than ADAM (Dozat, 2016; Géron, 2022). For the proposed model, we use the NADAM optimizer.

Early stopping criteria with a model checkpoint are implemented to find an optimal number of epochs. We take a sufficiently large number as epochs and monitor the loss. If the loss does not decrease for the last five epochs, then the model stops running, and predictive performances are ready to generate. Early stopping and loss monitoring criteria also address the consequence of over-fitting when training LSTM, GRU, and hybrid GRNN models.

**Table 3.** TPE Bayesian optimization HPO results for Single-layer GRNN models.

| Hyperparameter | Alternatives | Model Architectures | Optimum hyperparameters obtained from TPE Bayesian optimization HPO |
|---|---|---|---|
| Number of Neurons or Units in each layer | [32, 512] | LSTM | 47 |
| | [32, 512] | GRU | 95 |
| | [32, 512] | GRU-LSTM | (498, 311) |
| | [32, 512] | LSTM-GRU | (164, 52) |
| Learning Rate | $[10^{-4}, 10^{-2}]$ | LSTM | 0.00486 |
| | | GRU | 0.00016 |
| | | GRU-LSTM | 0.00019 |
| | | LSTM-GRU | 0.0004 |
| Batch Size | [16, 128] | LSTM | 46 |
| | [16, 128] | GRU | 54 |
| | [16, 128] | GRU-LSTM | 46 |
| | [16, 128] | LSTM-GRU | 66 |

**Table 4.** Hyperparameters constant for each GRNN model

| Hyperparameter | Chosen Parameter |
|---|---|
| Activation function for hidden layers | ReLU |
| Optimizer | Nadam |
| Number of epochs | Automatically obtained from loss monitoring and early stopping criteria with patience=5 |
| Loss function | MSE (mean square error) |

**Table 5.** TPE Bayesian optimization HPO results for Double-layer GRNN models.

| Hyperparameter | Alternatives | Model Architectures | Optimum hyperparameters obtained from TPE Bayesian optimization HPO |
|---|---|---|---|
| Number of Neurons or Units in each layer | [32, 512] | LSTM | 357, 250 |
| | [32, 512] | GRU | 394, 451 |
| | [32, 512] | GRU-LSTM | (437, 252), (68, 511) |
| | [32, 512] | LSTM-GRU | (230, 132), (108, 105) |
| Learning Rate | $[10^{-4}, 10^{-2}]$ | LSTM | 0.0014 |
| | | GRU | 0.00992 |
| | | GRU-LSTM | 0.0001 |
| | | LSTM-GRU | 0.00034 |
| Batch Size | [16, 128] | LSTM | 22 |
| | [16, 128] | GRU | 37 |
| | [16, 128] | GRU-LSTM | 102 |
| | [16, 128] | LSTM-GRU | 22 |

**Table 6.** TPE Bayesian optimization HPO results for Triple-layer GRNN models.

| Hyperparameter | Alternatives | Model Architectures | Optimum hyperparameters obtained from TPE Bayesian optimization HPO |
|---|---|---|---|
| Number of Neurons or Units in each layer | [32, 512] | LSTM | 168, 484, 71 |
| | [32, 512] | GRU | 230, 196, 273 |
| | [32, 512] | GRU-LSTM | (114, 101), (184, 217), (500, 229) |
| | [32, 512] | LSTM-GRU | (176, 126), (430, 146), (32, 83) |
| Learning Rate | $[10^{-4}, 10^{-2}]$ | LSTM | 0.0002 |
| | | GRU | 0.00887 |
| | | GRU-LSTM | 0.00016 |
| | | LSTM-GRU | 0.00055 |
| Batch Size | [16, 128] | LSTM | 98 |
| | [16, 128] | GRU | 42 |
| | [16, 128] | GRU-LSTM | 22 |
| | [16, 128] | LSTM-GRU | 44 |

### 5.4 Experimental results

The optimum hyperparameters for each GRNN model are found using TPE Bayesian optimization for HPO, presented in Table 3, Table 4 (for single layer), Table 5, and Table 6 (for multi-layer). Single and multilayer GRNN models (LSTM, GRU, GRU-LSTM, LSTM-GRU) are trained. The models are then fitted into the test data to assess the predictive capability of each model.

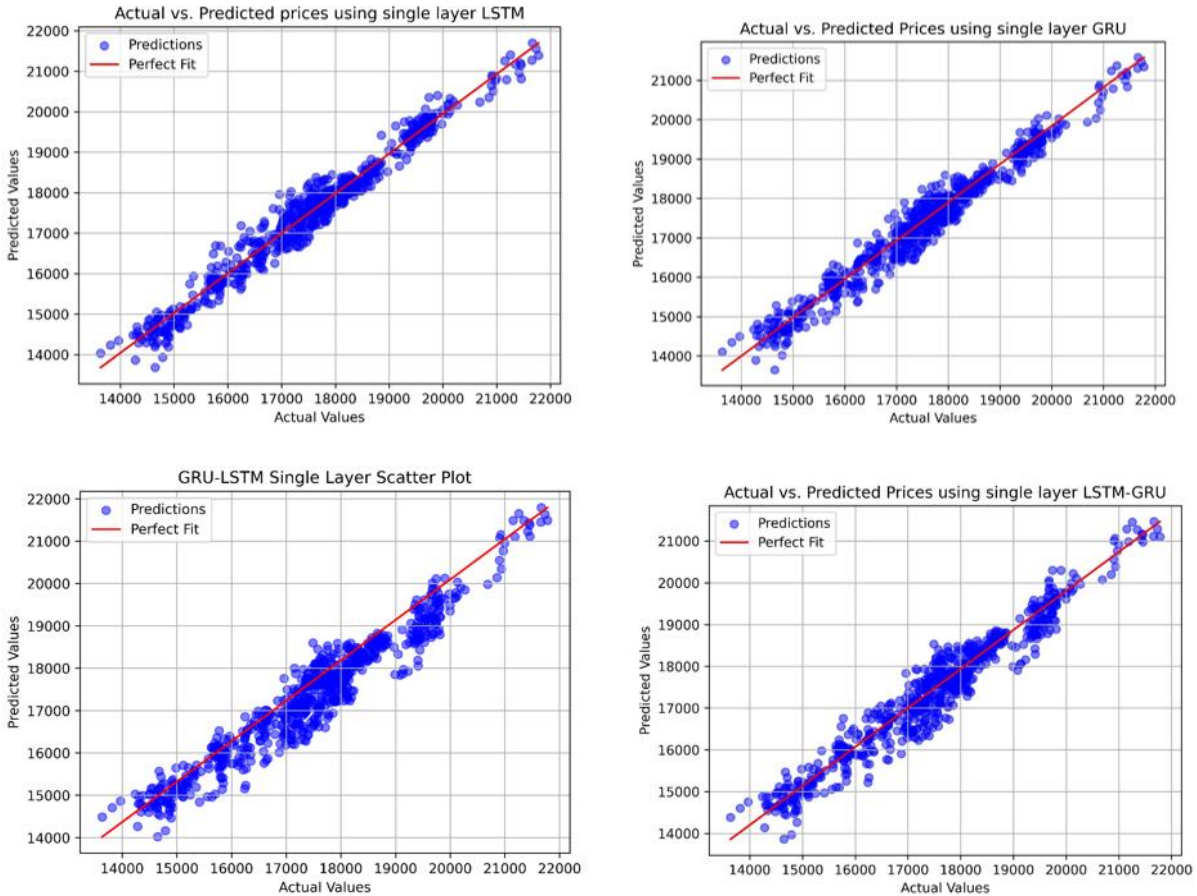

**Fig. 5.** Scatter plot of Predicted vs. Actual prices on test set under single-layer models: LSTM, GRU, GRU-LSTM, and LSTM-GRU

To resolve the stochastic nature of deep learning models, we have trained each model at least 48 times. We saved each model with an $R^2$ score greater than 0.90 as .keras or .h5 file. We choose the model with the best $R^2$ score for each GRNN architecture among the saved models. Then, we plot the models for each GRNN architecture. The performance of single and multi-layer GRNN models is presented in Table 7, Fig. 5, and Fig. 6.

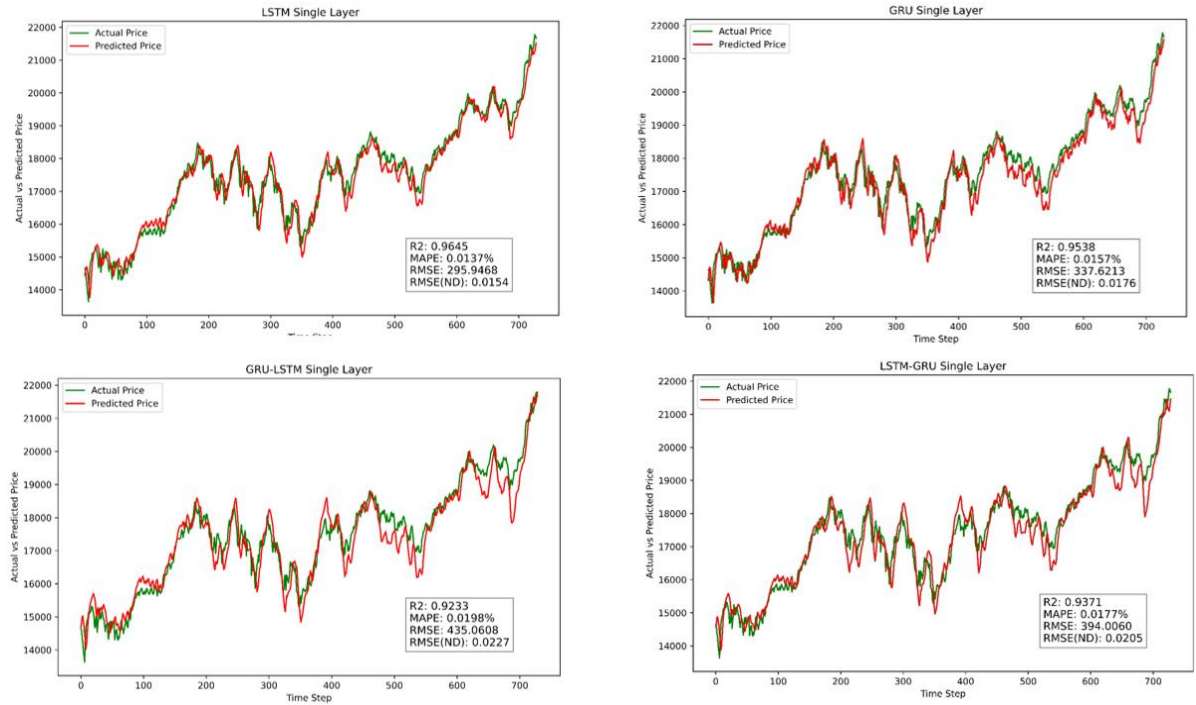

**Fig. 6.** Predicted vs. Actual prices on test set under single-layer GRNN models

**Table 7**. Performance table for single-layer TPE-GRNN models

| Models | Layers | $R^2$ | MAPE (%) | RMSE | RMSE (ND) |
|---|---|---|---|---|---|
| LSTM | 1 | **0.9645** | **0.0137** | **295.9468** | **0.0154** |
| GRU | 1 | 0.9538 | 0.0157 | 337.6213 | 0.0176 |
| GRU-LSTM | 1 | 0.9233 | 0.0198 | 435.0608 | 0.0227 |
| LSTM-GRU | 1 | 0.9371 | 0.0177 | 394.0060 | 0.0205 |

It appears that single-layer TPE-LSTM performs better than other considered single-layer TPE-GRNN models for the considered dataset.

Table 8 shows the prediction accuracy of double-layer TPE-GRNN models with optimum hyperparameters. The double-layer TPE-GRU model appears to represent better prediction accuracy than other double-layer TPE-GRNN models. Fig. 7 presents actual vs. predicted prices on test data of NIFTY 50 for double-layer TPE-GRNN models.

Table 9 shows the predictive performance of triple-layer TPE-GRNN models with optimum hyperparameters. The triple-layer TPE-LSTM model appears to represent better prediction accuracy among triple-layer TPE-GRNN models. Fig. 8 presents actual vs predicted prices on test data of NIFTY 50 for triple-layer TPE-GRNN models.

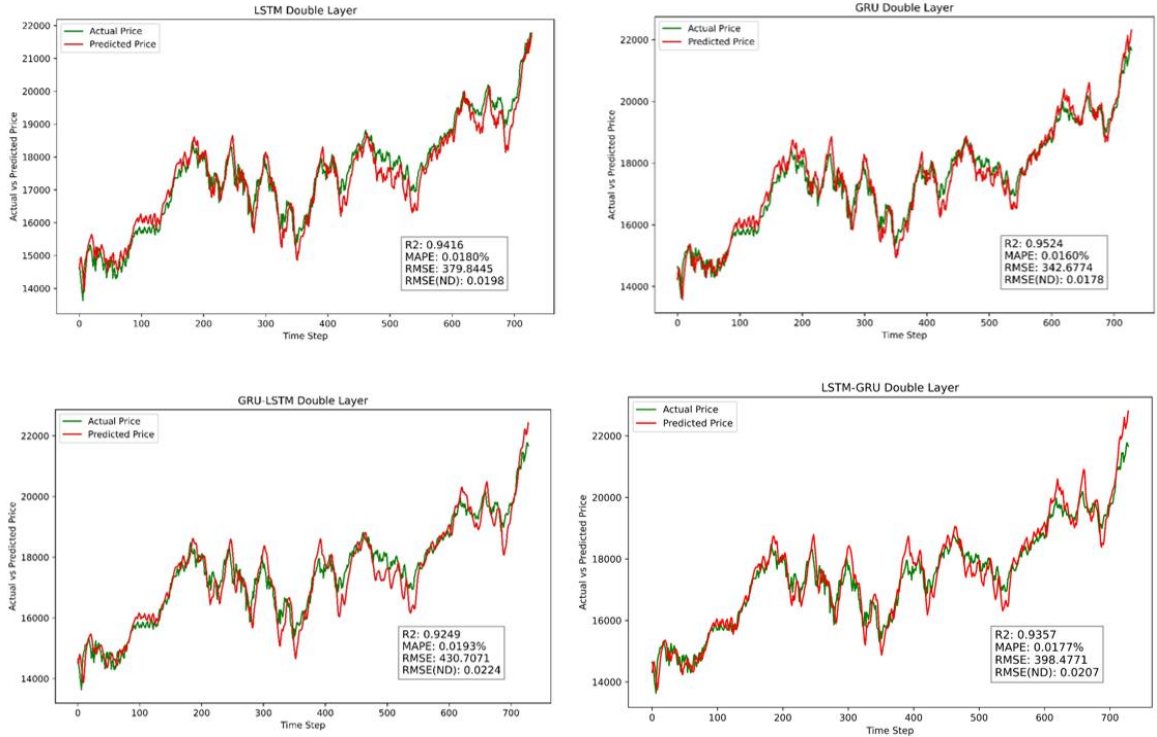

**Fig. 7.** Predicted vs. Actual prices on test set under double-layer GRNN models

**Table 8**. Performance table for double-layer TPE-GRNN models

| Models | Layers | $R^2$ | MAPE (%) | RMSE | RMSE (ND) |
|---|---|---|---|---|---|
| LSTM | 2 | 0.9416 | 0.0180 | 379.8445 | 0.0198 |
| GRU | 2 | **0.9524** | **0.0160** | **342.6774** | **0.0178** |
| GRU-LSTM | 2 | 0.9249 | 0.0193 | 430.7071 | 0.0224 |
| LSTM-GRU | 2 | 0.9357 | 0.0177 | 398.4771 | 0.0207 |

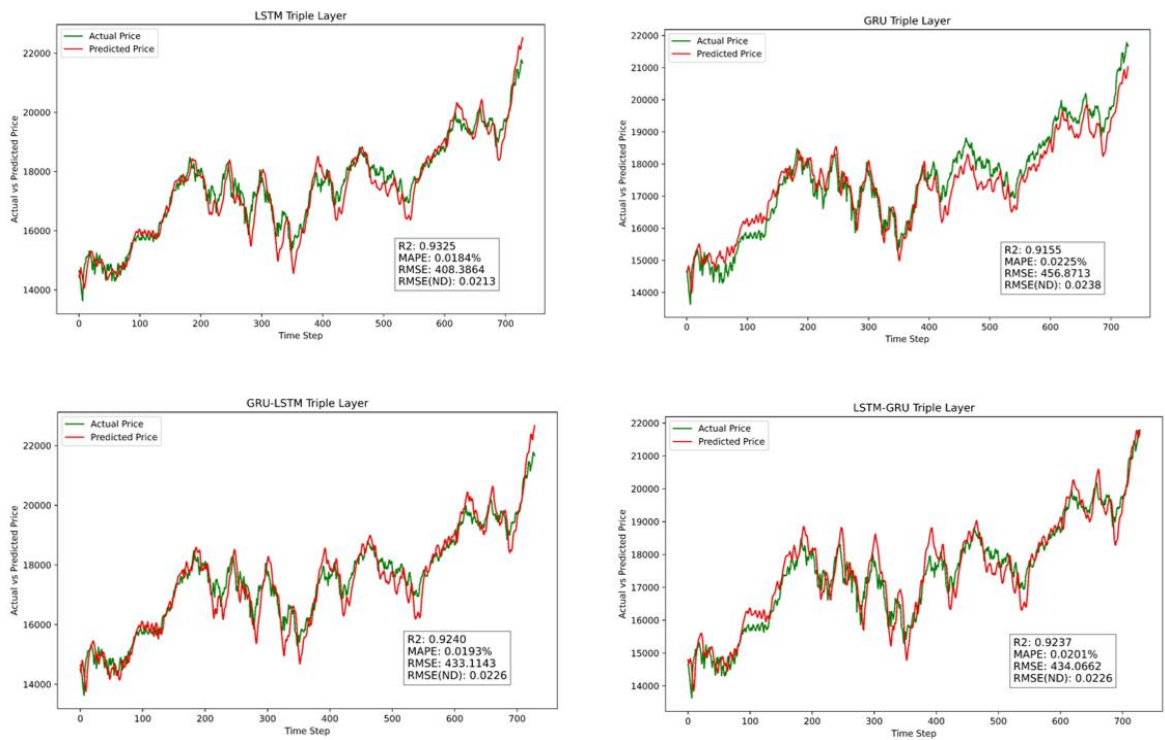

**Fig. 8.** Predicted vs. Actual prices on test set under triple-layer GRNN models

Table 9. Performance table for triple layer TPE-GRNN models

| Models | Layers | $R^2$ | MAPE (%) | RMSE | RMSE (ND) |
|---|---|---|---|---|---|
| LSTM | 3 | **0.9325** | **0.0184** | **408.3864** | **0.0213** |
| GRU | 3 | 0.9155 | 0.0225 | 456.8713 | 0.0238 |
| GRU-LSTM | 3 | 0.9240 | 0.0193 | 433.1143 | 0.0226 |
| LSTM-GRU | 3 | 0.9237 | 0.0201 | 434.0662 | 0.0226 |

It is observed that the evaluation matrices for single-layer TPE-GRNN models performs better than the multilayer TPE-GRNN models. Comparing all the considered TPE-GRNN models' performances, it is observed that a single-layer TPE-LSTM model gives the best prediction accuracy for forecasting the next-day closing price of the NIFTY 50 index.

5.5 *Prediction analysis and comparison*

To quantify the enhanced prediction accuracy of the proposed TPE-GRNN models more evidently, the rate of the decrease of MAPE is presented. MAPE for comparison is chosen because the stock return is calculated by percentage, and the scale of data does not change MAPE. The best MAPE of our proposed method (TPE-LSTM) is 0.0137%.

Table 10. MAPE comparison of previous models

| Year | Research Work | Dataset | Model Structure | HPO Method | Results | |
|---|---|---|---|---|---|---|
| | | | | | Best Model | MAPE (%) |
| 2017 | (Bao et al., 2017) | NIFTY 50 and other indices | RNN, LSTM, WLSTM, WSAEs-LSTM | - | WSAEs-LSTM | 0.024 |
| 2018 | (Hossain et al., 2018) | S&P 500 | LSTM-GRU | - | LSTM-GRU | 4.13 |
| 2019 | (Yu & Yan, 2019) | S&P 500 and other indices | ARIMA, SVR, MLP, LSTM | - | LSTM | 1.30 |
| 2022 | (Pokhrel et al., 2022) | NEPSE | LSTM, GRU, CNN | Manual Search | LSTM GRU | 0.6488 0.7350 |
| 2022 | (Bhandari et al., 2022) | S&P 500 | LSTM | Manual Search | LSTM | 0.7008 |
| 2022 | (Deng et al., 2022) | S&P 500, HSI, SSE | ARIMA, BPNN, LSTM, EMD-LSTM, MEMD-LSTM | OATM | MEMD-LSTM | 0.5276 (S&P 500) 0.5118(HIS) 0.3812(SSE) |
| 2022 | (Wang et al., 2022) | SSE, NIKKEI, KOSPI, SET | LSTM, VMD-LSTM, VMD-ALSTM, IVMD-ICEEMDAN-ALSTM, IVMD-ICEEMDAN-MFA-ALSTM, IVMD-ICEEMDAN-GRU, IVMD-ICEEMDAN-RNN, CEEMD-CNN-LSTM, Two-stage | | IVMD-ICEEMDAN-MFA-ALSTM | 0.612(SSE) 0.903(NIKKEI) 0.606(KOSPI) 0.402(SET) |
| | | | SVR-ANN, Random forest with technical parameters | | | |
| 2023 | (Gülmez, 2023) | DOW | LSTM-ARO, LSTM-GA, LSTM1D, LSTM2D, LSTM3D, ANN | ARO | LSTM-ARO | 0.020 |
| 2024 | (Beniwal et al., 2024) | NIFTY 50 | DNN, RNN, LSTM, Bi-LSTM, GRU, CNN | - | LSTM | 3.95 |
| 2024 | Proposed method | NIFTY 50 | LSTM, GRU, GRU-LSTM, LSTM-GRU | TPE Bayesian Optimization | TPE-LSTM | **0.0137** |

The MAPE of the proposed method has been improved by 42.92% compared to WSAEs-LSTM (Bao et al., 2017). The MAPE of the TPE-LSTM-GRU model of the proposed method has been improved by 99.57% compared to the LSTM-GRU implemented in (Hossain et al., 2018). Also, the MAPE of the proposed method has been improved by 98.95% and 99.65%, respectively, when compared to LSTM models proposed by (Yu & Yan, 2019) and (Beniwal et al., 2024). The MAPE of the proposed TPE-LSTM and TPE-GRU models are improved by 97.89% and 97.86% compared to LSTM and GRU models proposed in (Pokhrel et al., 2022), where the manual search for HPO is done. Also, The MAPE of the proposed method has been improved by 98.05% compared to the LSTM model proposed in (Bhandari et al., 2022), where the manual search for HPO is done.

The MAPE of the proposed method has been improved by 31.50% compared to the LSTM-ARO model (Gülmez, 2023), where the HPO method ARO is performed. The MAPE of the proposed method has been improved by 97.40% compared to one step-ahead prediction of the MEMD-LSTM method (Deng et al., 2022), where the HPO method OATM is applied. Furthermore, The MAPE of the proposed method has been improved by 96.59% when compared to the IVMD-ICEEMDAN-MFA-ALSTM method (Wang et al., 2022).

Table 10 and the comparison presented above show that proper HPO can significantly improve the model performance. Also, the appropriate choice of input factors can indicatively enhance the model's prediction accuracy.

## 6. Statistical test

A statistical test is conducted to validate the authenticity of the model's performance. First, we perform the normality test (D'Agostino-Pearson Test) for RMSE, MAPE, and R2 errors. The test results (Table 11) show that the values of RMSE, MAPE, and R2 errors of the best LSTM, GRU, GRU-LSTM, LSTM-GRU models of respective layers are in normal distribution since the p-values are higher than the significance level $\alpha = 0.05$.

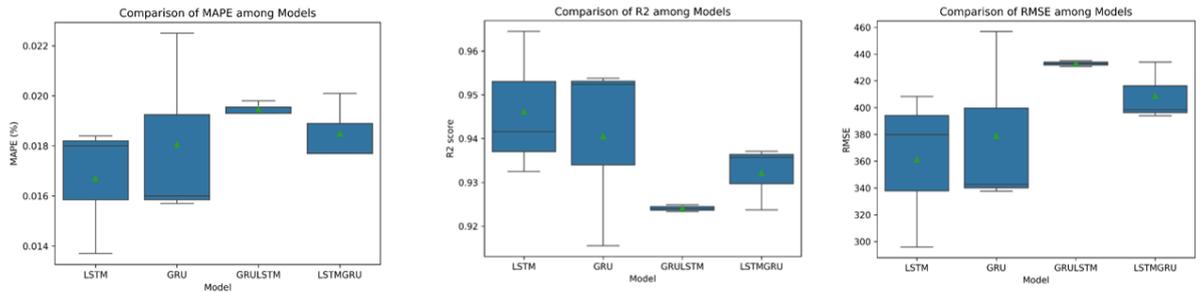

**Fig. 9.** The boxplot of error matrices of different models: RMSE, $R^2$, MAPE

**Table 11**. Normality test results of the errors

|  |  | LSTM | GRU | GRU-LSTM | LSTM-GRU |
|---|---|---|---|---|---|
| RMSE | t-statistic | 3.1913 | 3.8319 | 2.4298 | 3.7197 |
|  | p-value | 0.2028 | 0.1472 | 0.2967 | 0.1557 |
| $R^2$ | t-statistic | 3.0464 | 3.8366 | 2.4492 | 3.7362 |
|  | p-value | 0.2180 | 0.1469 | 0.2939 | 0.1544 |
| MAPE | t-statistic | 3.7752 | 3.8304 | 3.8500 | 3.8499 |
|  | p-value | 0.1514 | 0.1473 | 0.1459 | 0.1459 |

We perform two-sample t-tests (Welch's Test) to identify whether the models' prediction errors differ significantly. The p-value and the test statistic are presented in Table 12. The results show a substantial difference between the means of RMSE, MAPE, and $R^2$ under LSTM and GRU-LSTM models. The outcomes of the two sample t-tests indicate that LSTM is the best model among the models we considered for this dataset. Hence, the single-layer LSTM model gives the best prediction accuracy for forecasting the next day's closing price of NIFTY 50 index.

**Table 12**. Two sample t-test results

|  |  | (LSTM, GRU) | (LSTM, GRU-LSTM) | (LSTM, LSTM-GRU) | (GRU, GRU-LSTM) | (GRU, LSTM-GRU) | (LSTM-GRU, GRU-LSTM) |
|---|---|---|---|---|---|---|---|
| RMSE | t-statistic | -0.6857 | -4.2389 | -2.6332 | -2.7675 | -1.4553 | -3.7861 |
|  | p-value | 0.5029 | 0.0028 | 0.0246 | 0.0243 | 0.1773 | 0.0052 |
| MAPE | t-statistic | -1.0198 | -3.8759 | -2.1128 | -1.4085 | -0.3675 | -2.7738 |
|  | p-value | 0.3251 | 0.0045 | 0.0559 | 0.1962 | 0.7209 | 0.0223 |
| $R^2$ | t-statistic | 0.7156 | 4.1032 | 2.9247 | 2.2719 | 1.4405 | 2.1371 |
|  | p-value | 0.4853 | 0.0027 | 0.0141 | 0.0506 | 0.1814 | 0.0522 |

## 7. Conclusion

This research looked into improving the prediction accuracy of the NIFTY 50 index using gated RNN models with proper choice of input features, and TPE Bayesian optimization for HPO. The TPE-GRNN models are trained, tested, and compared to determine which model performs best for the NIFTY 50 index. It is found that the single-layer TPE-LSTM model (MAPE= 0.0137%, $R^2$ =0.9645, and RMSE(ND)=0.0154) provides superior prediction accuracy among single and multilayer TPE-GRNN models. It is shown that the MAPE of the proposed method is lowest (best) for stock index prediction, with respect to the other methods. The HPO and proper choice of input features significantly improve the models' prediction accuracy.

This model can also be applied to other broad market indexes if the data behaves similarly. Investors can use this model as an extra tool to make better trading decisions. However, the trading decision should not be based only on the outcome of this study. The investment decision should be made by considering other aspects of the economy and the risk tolerance ratio of an individual.

The future aim is to predict multi-step ahead stock price prediction and check if including unstructured textual data like social media sentiment, company earnings reports, breaking news on policies, and analyst research reports as features can improve the model's prediction accuracy. Additionally, there is interest in developing hybrid models that combine LSTM and GRU with other machine learning models. Finally, the future plan is to explore using hybrid neural network algorithms that combine local optimizers with global optimization techniques like artificial rabbit optimization, genetic algorithms, and particle swarm optimization.

**Data and Code availability**

The data used are publicly available (collected from Yahoo Finance and investing.com). The code and dataset are available at https://github.com/bvsdinda/TPE-LSTM.git